\pdfoutput=1
\documentclass[11pt]{article}
\usepackage{acl}
\usepackage{times}
\usepackage{latexsym}
\usepackage{array,xcolor}  % xcolor 早已需要
\newcolumntype{P}{@{}c@{\hspace{2pt}}c}  % 原值列 + 差值列，只隔 2pt

\usepackage[T1]{fontenc}
\usepackage[utf8]{inputenc}
\usepackage{microtype}
\usepackage{inconsolata}

\usepackage{amsthm}

\usepackage{amsmath}
\usepackage{graphicx}
\usepackage{algorithm}
\usepackage[noend]{algpseudocode}

\usepackage{enumitem}

\usepackage{multirow}
\usepackage{adjustbox}
\usepackage{bbm}
\usepackage{wrapfig,lipsum}
\usepackage{xspace}
\usepackage{booktabs,cellspace}  
\usepackage[normalem]{ulem}
\usepackage{makecell}
\usepackage{amssymb}% http://ctan.org/pkg/amssymb
\usepackage{pifont}% http://ctan.org/pkg/pifont
\usepackage{todonotes}
\usepackage{lipsum}
\usepackage{afterpage}

\usepackage{microtype}

\usepackage{cleveref}

\newcommand{\gpt}{\textsc{GPT-3.5-Turbo}\xspace}

\usepackage{etoolbox}
\usepackage[tikz]{bclogo}

\usepackage{subcaption}
\usepackage{adjustbox}

\definecolor{msftBlue}{RGB}{0,164,239}
\definecolor{msftGreen}{RGB}{127,186,0}
\definecolor{msftYello}{RGB}{255,185,0}
\definecolor{msftBlack}{RGB}{0,0,0}

\usepackage{tcolorbox} % for tcolorbox
\tcbuselibrary{skins} % for rounded corners
\usepackage[T1]{fontenc}

\tcbset{
    userstyle/.style={
        enhanced,
        colback=white,
        colframe=black,
        colbacktitle=gray!20,
        coltitle=black,
        rounded corners,
        sharp corners=north,
        boxrule=0.5pt,
        drop shadow=black!50!white,
        attach boxed title to top left={
            xshift=-2mm,
            yshift=-2mm
        },
        boxed title style={
            rounded corners,
            size=small,
            colback=gray!20
        }
    },
    replystyleg/.style={
        enhanced,
        colback=green!15,
        colframe=black,
        colbacktitle=green!30,
        coltitle=black,
        boxrule=0.5pt,
        drop shadow=black!50!white,
        rounded corners,
        sharp corners=north,
        attach boxed title to top right={
            xshift=-2mm,
            yshift=-2mm
        },
        boxed title style={
            rounded corners,
            size=small,
            colback=green!40
        }
    },
    replystyler/.style={
        enhanced,
        colback=red!15,
        colframe=black,
        colbacktitle=red!40,
        coltitle=black,
        boxrule=0.5pt,
        drop shadow=black!50!white,
        rounded corners,
        sharp corners=north,
        attach boxed title to top right={
            xshift=-2mm,
            yshift=-2mm
        },
        boxed title style={
            rounded corners,
            size=small,
            colback=red!40
        }
    }
}
\newtcolorbox{userquery}[1][]{
    userstyle,
    title=Probed Fuses for \gpt,
    #1
}

\newcommand{\mymethod}{\emph{MARCUS}}
\newcommand{\clean}{\emph{MARCUS-C}}
\newcommand{\splits}{\emph{MARCUS-S}}

\title{BRIGHT\textsuperscript{+}: Upgrading the BRIGHT Benchmark with \mymethod{}, a Multi-Agent RAG Clean-Up Suite}

\author{
Liyang Chen$^{\dagger}$ \quad 
Yujun Cai$^{\mathsection}$ \quad 
Jieqiong Dong$^{\ddagger}$ \quad 
Yiwei Wang$^{\mathparagraph}$ \\
$^\dagger$ University of California, Los Angeles \\
$^\mathsection$ The University of Queensland \\
$^\ddagger$ Central South University \\
$^\mathparagraph$ University of California, Merced \\
}

\date{}

\begin{document}
\maketitle
\begin{abstract}
Retrieval-Augmented Generation (RAG) systems require corpora that are both structurally clean and semantically coherent. \textbf{BRIGHT} is a recent and influential benchmark designed to evaluate complex multi-hop retrieval across diverse, high-reasoning domains. However, its practical effectiveness is limited by common web-crawled artifacts—such as content redundancy and semantic discontinuity—that impair retrieval accuracy and downstream reasoning. Notably, we find that such issues are concentrated in seven StackExchange-derived subdomains, while other domains (e.g., Coding and Theorem-based content) remain relatively clean.

In this study, we present \textbf{\mymethod{}}, a multi-agent pipeline that leverages large language models (LLMs) to systematically clean and re-chunk BRIGHT into a higher-quality corpus: \textbf{BRIGHT\textsuperscript{+}}. \mymethod{} applies dedicated agents for structural noise removal and semantic segmentation, preserving answer-bearing spans while improving contextual integrity. Experimental evaluations demonstrate that BRIGHT\textsuperscript{+} yields consistent and significant improvements in both retrieval accuracy and multi-hop reasoning across a diverse set of retrievers. We release both the BRIGHT\textsuperscript{+} corpus and the \mymethod{} pipeline to support future research on robust, reasoning-centric retrieval.
\end{abstract}

%%%%% Tanmay: no need to contend the content vs. position because the current writing make readers feel like position is more important than the content. but we're not actually trying to make that comparison. We just want to draw attention to the positional aspect and to say it's ALSO important. 

%%%%% Haikang: since there are only two positions you compared, it seems to be an overclaim to put the emphasize on position.

\begin{figure*}[t]
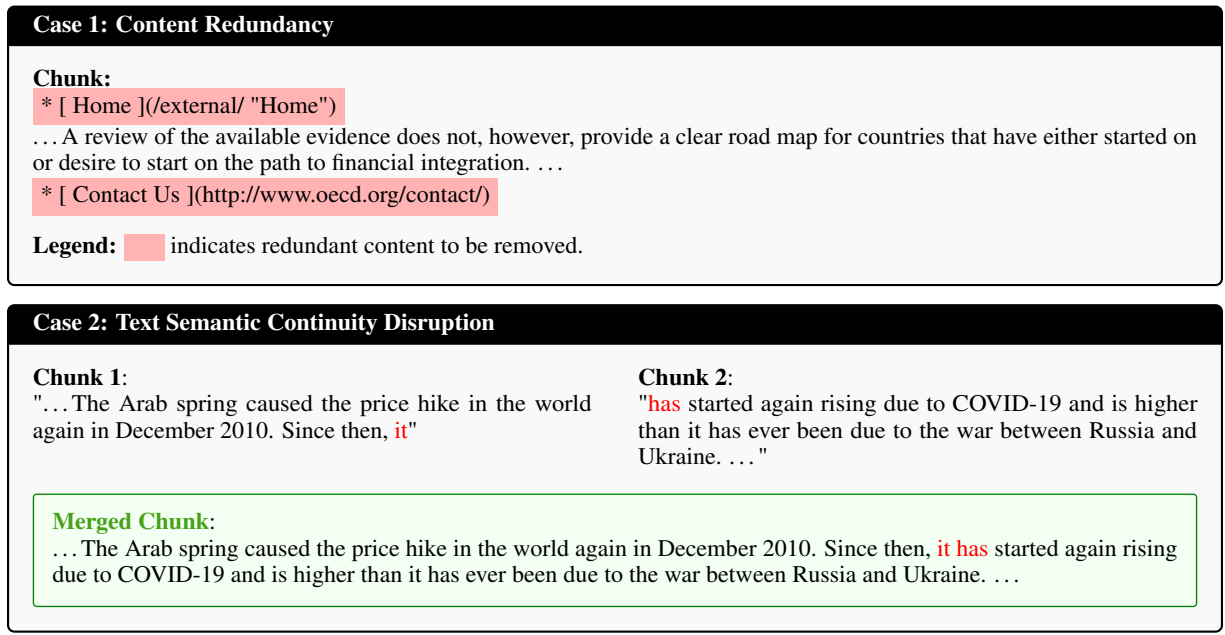

    \centering

    \begin{tcolorbox}[
        title=Case 1: Content Redundancy,
        coltitle=white,
        colback=gray!5!white,
        colframe=black,
        colbacktitle=black,
        boxrule=0.8pt,
        fonttitle=\bfseries\small,
        fontupper=\footnotesize,
        arc=2pt,
        width=\textwidth,
        left=6pt, right=6pt, top=6pt, bottom=6pt
    ]
    % {\color{red!30}\textbf{Chunk: }}
    {\textbf{Chunk: }}
    
    \colorbox{red!30}{* [ Home ](/external/ "Home")} \\
    \ldots A review of the available evidence does not, however, provide a clear road map for countries that have either started on or desire to start on the path to financial integration. \ldots \\
    \colorbox{red!30}{* [ Contact Us ](http://www.oecd.org/contact/)}
    
    \vspace{6pt}
    
    \noindent
    \textbf{Legend:} \colorbox{red!30}{\phantom{xx}} indicates redundant  content to be removed.
    \end{tcolorbox}

    \begin{tcolorbox}[
        title=Case 2: Text Semantic Continuity Disruption,
        coltitle=white,
        colback=gray!5!white,
        colframe=black,
        colbacktitle=black,
        boxrule=0.8pt,
        fonttitle=\bfseries\small,
        fontupper=\footnotesize,
        arc=2pt,
        width=\textwidth,
        left=6pt, right=6pt, top=6pt, bottom=6pt
    ]
    
    % 原始 Chunk 1 & 2
    \begin{minipage}[t]{0.48\textwidth}
    % {\color[RGB]{71,162,26}\textbf{Chunk 1}}:  
    {\textbf{Chunk 1}}:  
    
    "\ldots The Arab spring caused the price hike in the world again in December 2010. Since then, \textcolor{red}{it}"
    \end{minipage}%
    \hfill
    \begin{minipage}[t]{0.48\textwidth}
    \textbf{Chunk 2}:
    
    "\textcolor{red}{has} started again rising due to COVID-19 and is higher than it has ever been due to the war between Russia and Ukraine. \ldots"
    \end{minipage}

    \vspace{6pt}
    
        % Cleaned Chunk with visual highlight
        \begin{tcolorbox}[
            colback=green!5,
            colframe=green!50!black,
            boxrule=0.5pt,
            arc=1pt,
            left=4pt,
            right=4pt,
            top=4pt,
            bottom=4pt,
            enhanced,
            width=\textwidth
        ]
        {\color[RGB]{71,162,26}\textbf{Merged Chunk}}: \\
        \ldots The Arab spring caused the price hike in the world again in December 2010. Since then, \textcolor{red}{it has} started again rising due to COVID-19 and is higher than it has ever been due to the war between Russia and Ukraine. \ldots
        \end{tcolorbox}
    
    \end{tcolorbox}

    \caption{
    Illustration of two pervasive data quality issues in the original BRIGHT corpus that hinder retrieval-augmented reasoning: \textbf{(Top)} content-level redundancy introduced by web boilerplate; \textbf{(Bottom)} semantic discontinuity caused by coarse document segmentation. 
    These issues are not isolated cases but occur frequently and extensively throughout the corpus, motivating our LLM-based \mymethod{} pipeline for selective cleaning and coherent chunking.}

    \label{fig:bright-cases}
\end{figure*}

\section{Introduction}

Robust retrieval-augmented reasoning systems rely heavily on high-quality datasets—but real-world benchmarks often suffer from content-level redundancy introduced by web structures and disruptions to semantic continuity arising from coarse-grained document segmentation. These data quality issues are especially detrimental in retrieval-augmented reasoning tasks, where models must perform multi-hop inference and complex semantic analysis. Structural noise, excessive boilerplate, and fragmented semantic units force models to navigate through irrelevant content and arbitrary boundaries, ultimately compromising performance and distorting reasoning chains~\citep{jiang2020data, jiang2022closer}.

Against this backdrop, the BRIGHT dataset~\citep{BRIGHT} was introduced as a benchmark tailored for reasoning-intensive retrieval, aiming to advance beyond traditional benchmarks like BEIR~\citep{thakur2021beir} and MTEB~\citep{muennighoff2023mteb}. BRIGHT consists of three high-level domains—StackExchange, Coding, and Theorem-based content—spanning eleven subdomains in total. Notably, we find that serious quality issues are concentrated in seven StackExchange-derived subdomains, which exhibit pervasive structural artifacts typical of web-scraped data. These include noisy boilerplate, improperly segmented formulas and tables, and heavily fragmented semantic units that disrupt contextual understanding (see Figure~\ref{fig:bright-cases}). For a retriever, it's like reading a shuffled book filled with boilerplate—the narrative breaks down under noise and fragmentation, making reasoning harder than it should be.

Traditional data cleaning relies on manual annotation or rule-based heuristics—effective at small scales but costly, hard to scale, and ill-suited for capturing the semantic granularity required in reasoning tasks. Crucially, these methods are not designed for the demands of multi-hop inference, leaving a gap in scalable, semantically aware preprocessing tailored to retrieval-augmented reasoning.

To address these challenges, we introduce a framework that leverages LLMs to refine BRIGHT into a slimmer, more coherent corpus dubbed BRIGHT\textsuperscript{+}, tailored for retrieval-augmented reasoning.

Our contributions are threefold:
\begin{itemize}[leftmargin=1.25em, itemsep=0pt, parsep=0pt]
    \item We audit the benchmark and reveal two pervasive flaws, content-level redundancy and semantic discontinuity—localized to seven StackExchange subdomains—that jointly depress retrieval accuracy.
    \item We propose \mymethod{}, an LLM-driven pipeline whose dedicated cleaning and splitting agents adaptively prune noise and re-chunk documents into semantically coherent units.
    \item We release BRIGHT\textsuperscript{+}, a high-SNR version of BRIGHT to support retrieval research focused on model improvements rather than dataset noise.
\end{itemize}

\begin{figure*}[t]
    \centering
    \includegraphics[width=\textwidth]{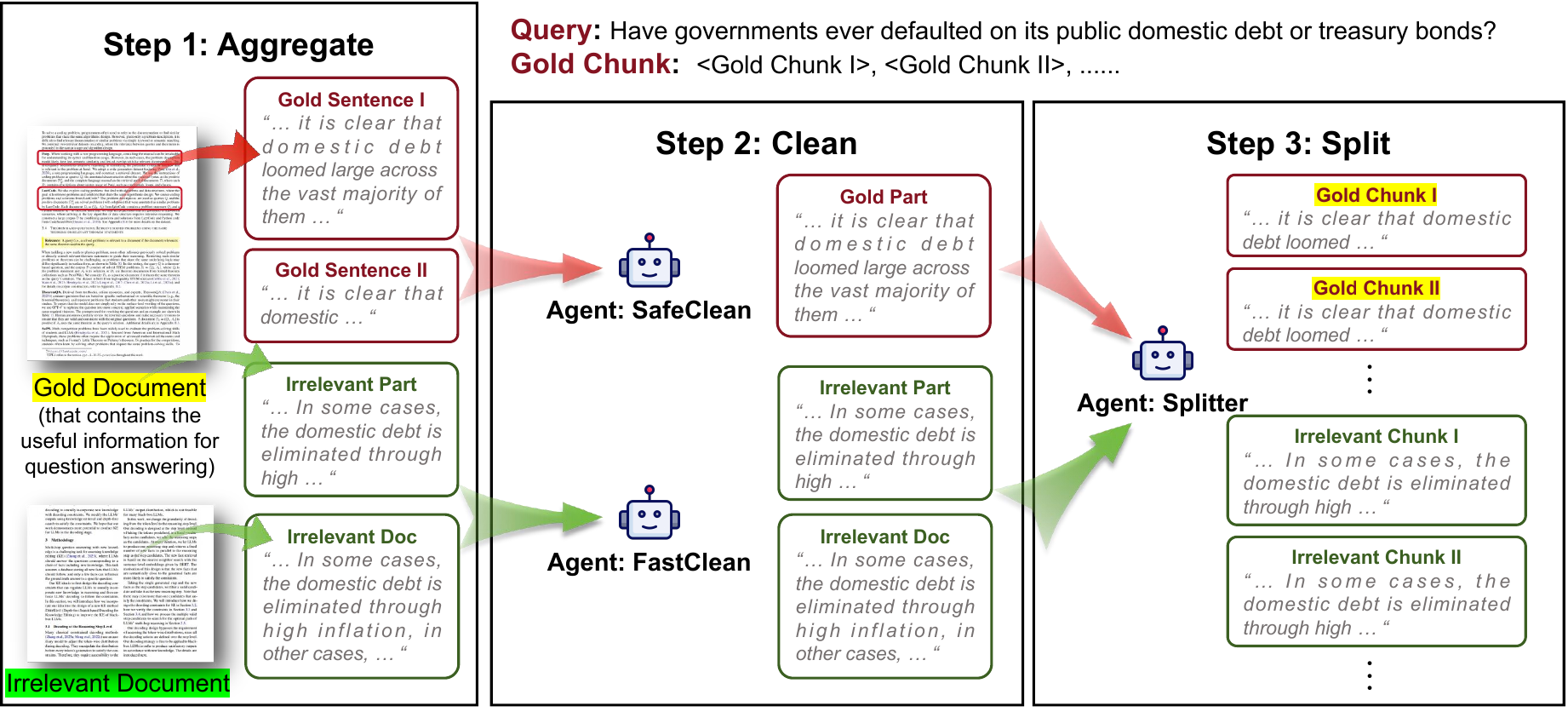}
    \caption{
    Overview of the \mymethod{} pipeline. BRIGHT documents are first split into gold and irrelevant parts based on span-level annotations. \textsc{SafeClean} and \textsc{FastClean} then remove structural noise with different strategies. Finally, \textsc{Splitter} segments the cleaned text into coherent chunks for retrieval-augmented reasoning.
    }
    \label{fig:pipeline}
\end{figure*}

\section{Related Work}

\paragraph{Retrieval Benchmarks.}
Retrieval benchmarks are central to evaluating the effectiveness and generalization of retrieval models. Standard datasets such as BEIR~\citep{thakur2021beir} and MTEB~\citep{muennighoff2023mteb} primarily test semantic similarity under relatively clean conditions, often overlooking challenges posed by noisy, fragmented, or web-derived content. These benchmarks typically emphasize single-hop factual retrieval, offering limited insight into a model’s reasoning capacity.

Recent efforts like BRIGHT~\citep{BRIGHT} attempt to address this by introducing reasoning-intensive tasks requiring multi-hop and domain-specific inference. However, BRIGHT inherits many issues common to web corpora—e.g., HTML artifacts, redundant navigational blocks, and poor segmentation—that disrupt semantic coherence and reduce benchmark reliability. This aligns with observations in prior work on web-crawled datasets~\citep{xu2017zipporah, khayrallah2018impact}, which highlight the prevalence of structural noise and redundant content.

\paragraph{Traditional Data Cleaning.}
To address such issues, early approaches have employed heuristic-based filtering, rule-based extraction, or manual annotation~\citep{grusky2018newsroom, nan2021entity, budach2022data}. While effective in limited contexts, these methods are costly, labor-intensive, and difficult to scale, especially for retrieval datasets where semantic coherence is key.

\paragraph{LLMs for Semantic Cleaning.}
LLMs offer a promising alternative for scalable and semantically-aware data cleaning. Leveraging deep contextual understanding, LLMs have been applied to tasks like text normalization, deduplication, and structural repair~\citep{sun2023chatgpt, lee2024gecko}. Recent work shows that LLM-derived embeddings can implicitly encode semantic and structural quality, aiding downstream retrieval performance. Despite these advances, most efforts focus on sentence-level annotation or re-ranking, with limited exploration of corpus-level cleaning tailored for retrieval-intensive reasoning tasks.

\begin{figure*}[t]
    \centering
    \includegraphics[width=\textwidth]{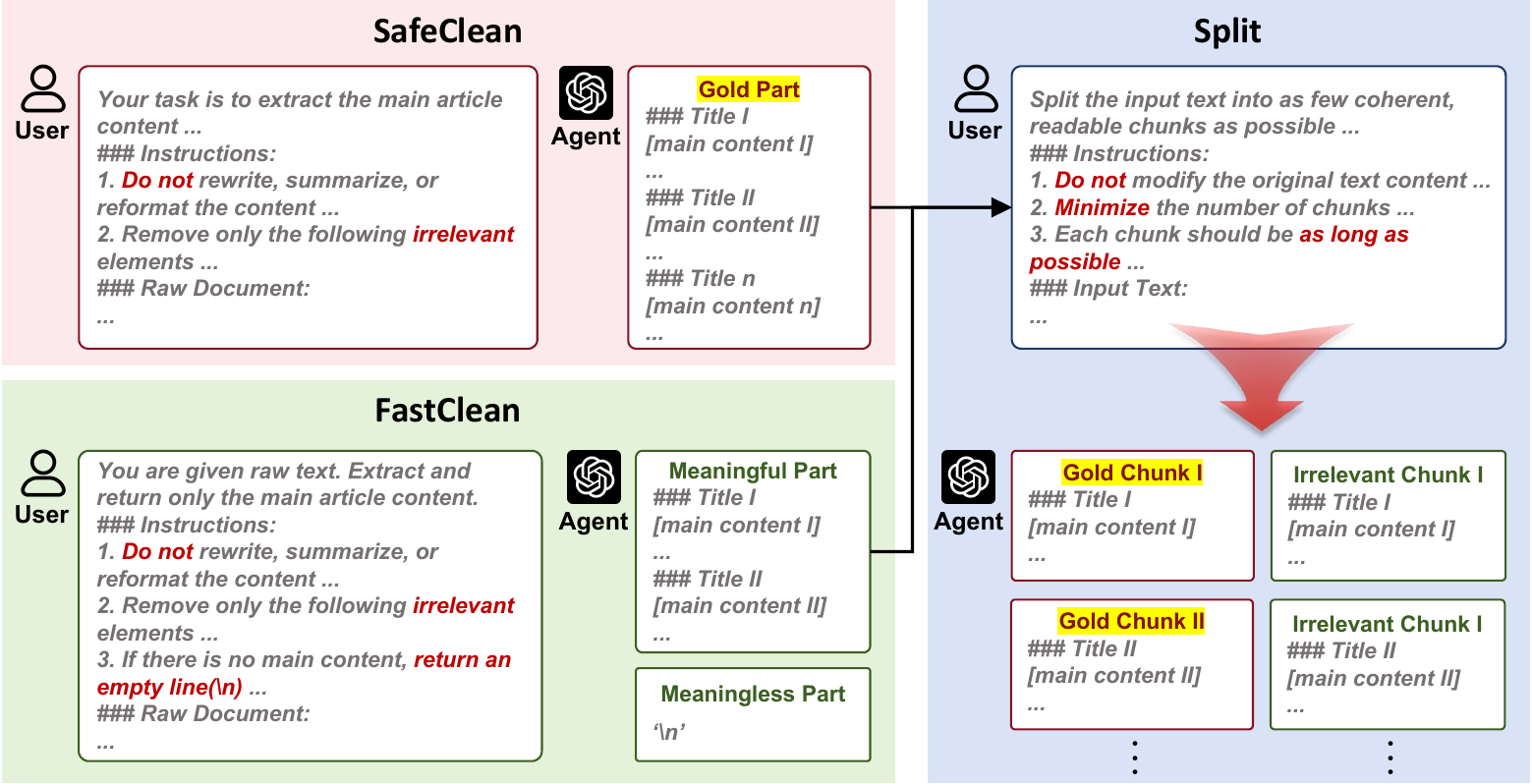}
    \caption{
    Prompt strategies for \mymethod{} agents: \textsc{SafeClean}, \textsc{FastClean}, and \textsc{Splitter}, each tailored for targeted filtering and semantic chunking. See Appendix~\ref{app:prompts} for full prompt templates.
    }
\label{fig:prompt}
\end{figure*}

\section{Methodology}
\label{sec:method}

We introduce \mymethod{}, a clean-and-split pipeline designed to address content-level redundancy and semantic discontinuity in the BRIGHT dataset.  
\mymethod{} leverages LLMs to systematically perform document cleaning and semantic segmentation, with special attention to preserving original annotations.

\vspace{-0.5em}
\subsection{Pipeline Overview}
As illustrated in Figure~\ref{fig:pipeline}, \mymethod{} operates in three sequential stages:

\begin{itemize}[leftmargin=1em,itemsep=1pt,topsep=2pt]
    \item \textbf{Step 1: Aggregate.}  
    The original long documents from the BRIGHT dataset are categorized into two types: \emph{Gold Documents}, which contain useful information relevant for answering at least one query from the benchmark, and \emph{Irrelevant Documents}, which do not contain any answer-supporting content. \emph{Gold Documents} are further decomposed into \emph{Gold Parts} (i.e., answer-containing spans) and \emph{Irrelevant Parts} (i.e., noisy or off-topic segments).

    \item \textbf{Step 2: Clean.}  
    We apply two dedicated LLM agents for content cleaning. \textbf{SafeClean} processes the Gold Parts to preserve their semantic integrity while removing structural noise. \textbf{FastClean} is applied to the Irrelevant Parts of Gold Documents and to all Irrelevant Documents, aggressively removing uninformative content. Both agents return fully reconstructed documents after cleaning.

    \item \textbf{Step 3: Split.}  
    All cleaned documents are then passed to \textbf{Splitter}, which performs LLM-based semantic segmentation. This step converts the documents into well-formed and semantically coherent chunks suitable for retrieval.
\end{itemize}
Figure~\ref{fig:prompt} illustrates the prompt configurations used by agents in the clean and split stages.

\subsection{Aggregate -- Gold vs. Irrelevant}
To initiate the cleaning process, we first decompose each gold document in the BRIGHT corpus into two disjoint regions: the \emph{Gold Parts} and the \emph{Irrelevant Parts}.  
The BRIGHT benchmark includes gold chunk annotations for each query, along with their relative positions within the source documents.  
By aggregating these gold chunks in their original order, we reconstruct the \emph{Gold Parts} of a document—that is, the span containing answer-supporting content.  
All remaining content in the same document is grouped as the \emph{Irrelevant Parts}, typically consisting of off-topic or noisy segments.

This decomposition strategy is tailored to BRIGHT but can generalize to other datasets that provide query-document alignment or rationale-level supervision.

\subsection{Clean -- Dual-Agent Noise Removal}
\subsubsection{SafeClean for Gold Parts}
\emph{Gold Parts} demand a conservative cleaning strategy, implemented by our \textbf{SafeClean} agent, that preserves every annotated span verbatim while excising only interface-level clutter. 
Accordingly, we prompt the LLM to perform selective deletion headers, sidebars, navigation widgets, edit buttons, donation banners, and similar artefacts are removed, whereas paragraph text, section headings, lists, and in-line citations are left untouched.  
The agent must output valid Markdown that replicates the original rhetorical structure without paraphrasing or summarising, thereby safeguarding the positional cues on which evaluation metrics rely.  
When encountering garbled characters, it attempts context-driven repair but leaves ambiguous cases unchanged to avoid hallucinated edits.  
This carefully constrained procedure eliminates noise that would mislead retrievers while guaranteeing that the semantic and positional integrity of the human annotations remains intact.

\subsubsection{FastClean for Irrelevant Contents}
In contrast to the \emph{Gold Parts}, the \emph{Irrelevant Parts} or \emph{Irrelevant Documents} of BRIGHT—comprising boilerplate, template fragments, or genuinely off-topic passages—pose no risk of annotation loss but do inflate the corpus with distracting noise.  
We therefore employ \textbf{FastClean} agent, an LLM agent configured with a fail-fast policy: it is instructed to excise the full gamut of interface artefacts (navigation bars, banners, login links, and similar chrome) and to return an empty string whenever no substantive article text remains.  
The prompt expressly forbids rewriting or reformatting; any residual content must be delivered verbatim as plain text.  
This aggressive strategy prioritises precision over recall—irrelevant tokens are discarded even at the cost of occasional under-retention—because these segments neither support evaluation nor influence ground-truth alignment.

\subsection{Split -- Semantic Chunking via LLM}
After cleaning, each document is passed through LLM-based \textbf{Splitter} agent that converts continuous text into a sequence of semantically self‑contained chunks.  
The agent operates under two complementary objectives: (\emph{i}) maximise intra–chunk coherence, and (\emph{ii}) minimise the number of chunks.  
In practice, chunk boundaries are introduced only at pronounced topic or discourse shifts, ensuring that each resulting unit preserves the local rhetorical flow.  
Structured elements such as tables or code snippets—when present—are rendered into concise natural‑language descriptions and incorporated within the nearest chunk, thereby maintaining information completeness without sacrificing textual uniformity.  
Each chunk is emitted in canonical order and assigned a deterministic label (\textsc{Chunk~A}, \textsc{Chunk~B}, …), yielding a discretised representation that is both human‑readable and directly consumable by retrieval models.  
This procedure reduces document fragmentation while retaining essential context, striking a favourable balance between retrieval granularity and semantic integrity.

\begin{table*}[t]
\centering
\caption{
Retrieval performance (nDCG@10, \%) on BRIGHT and BRIGHT\textsuperscript{+}. Right-hand columns indicate relative changes (\textcolor{red}{$\uparrow$} = improvement, \textcolor{blue}{$\downarrow$} = drop) from BRIGHT to BRIGHT\textsuperscript{+}.
}

\label{tab:bright-vs-brightplus}
\footnotesize
\begin{adjustbox}{width=1.0\linewidth}
\begin{tabular}{l*{8}{c c}}
\toprule
\textbf{Retriever} &
\multicolumn{2}{l}{Biology} &
\multicolumn{2}{l}{Earth Science} &
\multicolumn{2}{l}{Economics} &
\multicolumn{2}{l}{Psychology} &
\multicolumn{2}{l}{Robotics} &
\multicolumn{2}{l}{Stackoverflow} &
\multicolumn{2}{l}{Sustainable living} &
\multicolumn{2}{l}{Average}\\
\midrule
\multicolumn{17}{c}{\textbf{BRIGHT}}\\
\midrule
BM25   & 18.9 & -- & 27.2 & -- & 14.9 & -- & 12.5 & -- & 13.6 & -- & 18.4 & -- & 15.0 & -- & 17.2 & -- \\
BGE    & 11.7 & -- & 24.6 & -- & 16.6 & -- & 17.5 & -- & 11.7 & -- & 10.8 & -- & 13.3 & -- & 15.2 & -- \\
GritLM & 24.8 & -- & 32.3 & -- & 18.9 & -- & 19.8 & -- & 17.1 & -- & 13.6 & -- & 17.8 & -- & 20.6 & -- \\
QWEN2   & 30.6 & -- & 36.4 & -- & 17.8 & -- & 24.6 & -- & 13.2 & -- & 22.2 & -- & 14.8 & -- & 22.8 & -- \\
\midrule
\multicolumn{17}{c}{\textbf{BRIGHT\textsuperscript{+} (Ours)}}\\
\midrule
BM25   & 22.7 & {\color{red}$\uparrow$ 3.8} & 22.7 & {\color{blue}$\downarrow$ 4.5} & 16.1 & {\color{red}$\uparrow$ 1.2} & 12.7 & {\color{red}$\uparrow$ 0.2} & 21.7 & {\color{red}$\uparrow$ 8.1} & 11.6 & {\color{blue}$\downarrow$ 6.8} & 16.4 & {\color{red}$\uparrow$ 1.4} & 17.7 & {\color{red}$\uparrow$ 0.5} \\[2pt]
BGE    & 36.0 & {\color{red}$\uparrow$ 24.3} & 33.6 & {\color{red}$\uparrow$ 9.0}  & 31.0 & {\color{red}$\uparrow$ 14.4} & 25.4 & {\color{red}$\uparrow$ 7.9} & 27.2 & {\color{red}$\uparrow$ 15.5} & 32.9 & {\color{red}$\uparrow$ 22.1} & 34.7 & {\color{red}$\uparrow$ 21.4} & 31.5 & {\color{red}$\uparrow$ 16.3} \\[2pt]
GritLM & \underline{53.2} & {\color{red}$\uparrow$ 28.4} & \underline{43.3} & {\color{red}$\uparrow$ 11.0} & \textbf{40.2} & {\color{red}$\uparrow$ 21.3} & \underline{34.2} & {\color{red}$\uparrow$ 14.4} & \textbf{33.8} & {\color{red}$\uparrow$ 16.7} & \underline{36.4} & {\color{red}$\uparrow$ 22.8} & \underline{47.9} & {\color{red}$\uparrow$ 30.1} & \underline{41.3} & {\color{red}$\uparrow$ 20.7} \\[2pt]
QWEN2   & \textbf{62.2} & {\color{red}$\uparrow$ 31.6} & \textbf{45.6} & {\color{red}$\uparrow$ 9.2}  & \underline{37.1} & {\color{red}$\uparrow$ 19.3} & \textbf{36.8} & {\color{red}$\uparrow$ 12.2} & \underline{30.9} & {\color{red}$\uparrow$ 17.7} & \textbf{36.4} & {\color{red}$\uparrow$ 14.2} & \textbf{49.5} & {\color{red}$\uparrow$ 34.7} & \textbf{42.6} & {\color{red}$\uparrow$ 19.8} \\
\bottomrule
\end{tabular}
\end{adjustbox}
\end{table*}

\section{Experiments}
We conduct five experiments to evaluate our approach: retrieval benchmarking, corpus statistics analysis, ablation of pipeline components, reasoning performance with rewritten queries, and reranking effectiveness across retrievers.

\subsection{Experimental Setup}
\paragraph{Retrievers.}
We evaluate four representative retrievers spanning both sparse and dense paradigms. \textbf{BM25}~\citep{robertson2009probabilistic} is a classic sparse model based on lexical term overlap and serves as a strong non-neural baseline. \textbf{BGE} (335M;~\citealp{xiao2023BGE}) is a compact bilingual dense retriever optimized for sentence-level semantic matching. \textbf{GritLM} (7B;~\citealp{muennighoff2024gritlm}) is a large multilingual model trained on diverse retrieval tasks, exhibiting strong cross-lingual generalization. \textbf{QWEN2} (7B;~\citealp{qwen2}) is an instruction-tuned embedding model designed for high-performance dense retrieval, particularly effective on both English and Chinese corpora.

\paragraph{Metrics.}
We adopt two types of evaluation metrics in our experiments. For retrieval performance, we use \textbf{nDCG@10} as the primary metric, as it reflects both the relevance and the ranking quality of retrieved documents. In addition, for our ablation studies, we employ \textbf{LLM-based evaluation} where LLMs are used as both generators and evaluators to assess the quality of generated answers in a full RAG pipeline.

\subsection{Retrieval Performance}
The experimental results in Table~\ref{tab:bright-vs-brightplus} lead to several key findings:

\paragraph{BRIGHT+ substantially improves overall retrieval effectiveness.}
Dense neural retrievers (BGE, GritLM, QWEN2) register marked gains on every domain when evaluated on BRIGHT\textsuperscript{+}, whereas the sparse BM25 baseline shows only a modest average lift and even declines in a few cases (e.g., \emph{Earth Science}, \emph{Stackoverflow}).  
These results indicate that the cleaned and semantically re-structured corpus particularly amplifies the strengths of embedding-based methods, while adding little benefit—and occasional noise—to purely lexical ranking.  
By delivering a markedly higher signal-to-noise ratio, BRIGHT\textsuperscript{+} provides a more reliable test bed that lets researchers concentrate on retrieval modelling rather than dataset artefacts.

\paragraph{Performance gains are domain-dependent.} The most substantial improvements are found in \emph{Biology} and \emph{Sustainable.} (both over 30 pp), implying that documents in these fields particularly benefit from enhanced terminology standardization and contextual enrichment. On the other hand, BM25 performs worse in \emph{Earth.} and \emph{Stack.}, possibly due to detrimental shifts in term distribution caused by data rewriting.

\subsection{Corpus Statistics}
\paragraph{Chunk Reduction Statistics.}
A key objective of \mymethod{} is to mitigate structural redundancy and remove fragmented or noisy textual segments.  
Table~\ref{tab:chunk_count} quantifies the impact: the total chunk count drops from 511\,497 to 18\,187 (a \emph{97\%} reduction, retaining only 3.56\% of the original segments).  

\begin{table}[h]
    \centering
    \caption{Chunk count in BRIGHT and BRIGHT\textsuperscript{+}, with the retention ratio after cleaning.}
    \label{tab:chunk_count}
    \resizebox{\linewidth}{!}{
    \begin{tabular}{lccc}
        \toprule
        Sub-dataset & BRIGHT & BRIGHT\textsuperscript{+} (Ours) & Retention (\%) \\
        \midrule
        Biology   & 57\,359 & 2\,102 & 3.66 \\
        Earth. & 121\,249 & 4\,321 & 3.56 \\
        Econ.  & 50\,220 & 1\,634 & 3.25 \\
        Psy.   & 52\,835 & 2\,285 & 4.32 \\
        Rob.   & 61\,961 & 2\,385 & 3.85 \\
        Stack. & 107\,081 & 3\,894 & 3.64 \\
        Sus.   & 60\,792 & 1\,566 & 2.58 \\
        \midrule
        \textbf{Total} & \textbf{511\,497} & \textbf{18\,187} & \textbf{3.56} \\
        \bottomrule
    \end{tabular}
    }
\end{table}

\begin{figure}[h]
    \centering
    \includegraphics[width=\linewidth]{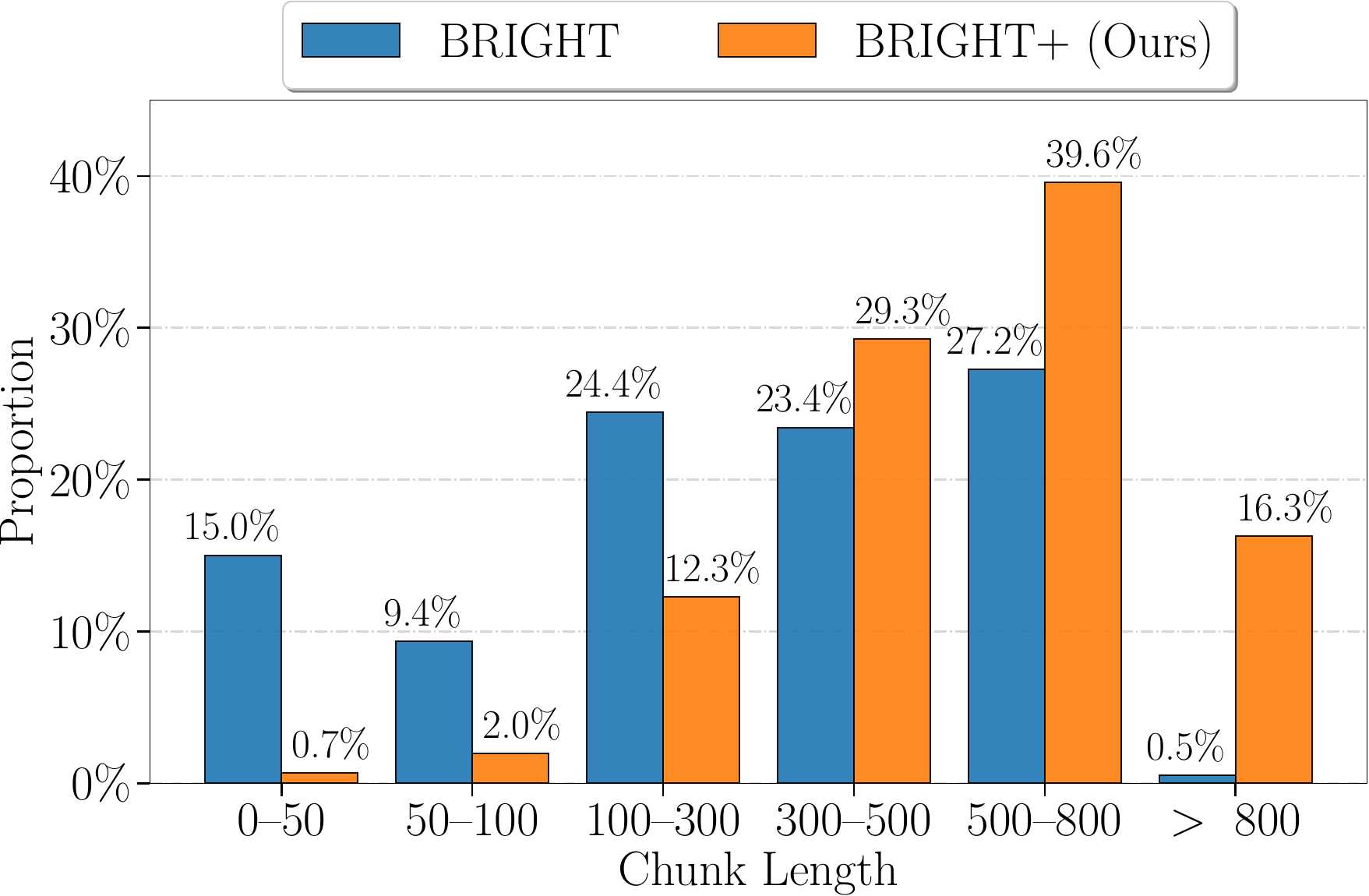}
    \caption{Distribution of chunk length in BRIGHT and BRIGHT\textsuperscript{+}.}
    \label{fig:chunk_length}
\end{figure}

\paragraph{Chunk Length Distribution.}
Figure~\ref{fig:chunk_length} shows that \mymethod{} removes most sub-100 token chunks (from $\sim$ 24\% to $<3\%$) and shifts the bulk of the corpus into the 300 to 800 token range. The resulting longer, self-contained chunks offer richer context windows, which help account for the observed improvements in retrieval performance.

\begin{figure}[h]
    \centering
    \includegraphics[width=\linewidth]{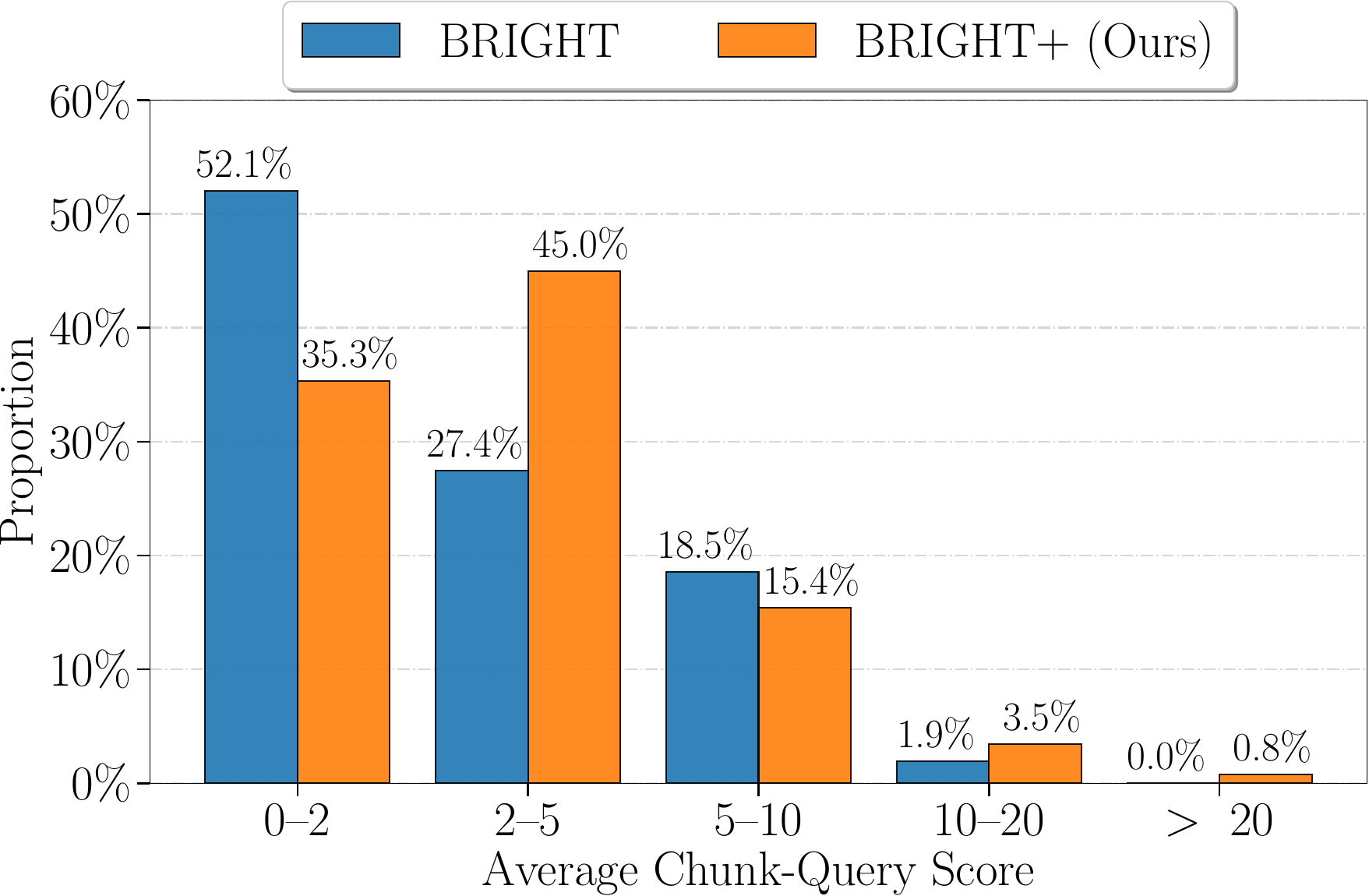}
    \caption{Distribution of chunk--query similarity scores in BRIGHT and BRIGHT\textsuperscript{+}.}
    \label{fig:score_dist}
\end{figure}

\paragraph{Average Chunk-Score Distribution.}
Figure~\ref{fig:score_dist} shows a clear shift toward higher query--chunk similarity scores after cleaning.  
Low-score matches become less dominant, while mid- and high-score ranges grow noticeably.  
This shift indicates that \mymethod{} reduces noise and improves the alignment between queries and relevant content, leading to better retrievability even for lexical models like BM25.

\subsection{Ablation Study}

To isolate the contributions of different components, we compare distinct cleaning strategies applied to the BRIGHT dataset. This allows us to better understand how each step—structural cleaning and semantic segmentation—affects downstream retrieval performance.

\begin{table*}[t]
\centering
\footnotesize
\begin{adjustbox}{width=1.0\linewidth}
\begin{tabular}{lcccccccc}
\toprule
\textbf{Method} & Biology & Earth Science & Economics & Psychology & Robotics & Stackoverflow & Sustainable Living & Average \\
\midrule
\multicolumn{9}{c}{\textbf{\clean{}}} \\
\midrule
BM25   & 12.5 & 15.9 &  9.4 &  5.1 &  8.1 &  6.1 & 10.5 &  9.7 \\
BGE    & 21.8 & 29.0 & 11.9 &  9.0 & 10.5 & 10.3 & 16.8 & 15.6 \\
GritLM & 22.5 & 14.6 &  7.0 &  5.8 &  5.3 &  2.4 & 13.5 & 10.1 \\
QWEN2  & 44.8 & 36.6 & 13.4 & 12.7 & 14.2 & 17.5 & 20.2 & 22.8 \\
\midrule
\multicolumn{9}{c}{\textbf{\splits{}}} \\
\midrule
BM25   & 20.3 & 19.1 & 18.4 & 15.9 & 20.7 & 12.6 & 24.0 & 18.7 \\
BGE    & 38.4 & 31.9 & 27.7 & 34.5 & 21.6 & 32.8 & 37.6 & 32.1 \\
GritLM & 54.2 & 40.0 & \underline{37.5} & \underline{47.7} & 31.2 & \textbf{36.7} & 48.7 & 42.3 \\
QWEN2  & \textbf{64.7} & 38.3 & 36.5 & \textbf{54.7} & \underline{31.8} & 35.2 & \textbf{55.0} & \textbf{45.2} \\
\midrule
\multicolumn{9}{c}{\textbf{\mymethod{} (Ours)}} \\
\midrule
BM25   & 22.7 & 22.7 & 16.1 & 12.7 & 21.7 & 11.6 & 16.4 & 17.7 \\
BGE    & 36.1 & 33.6 & 31.0 & 25.4 & 27.8 & 32.9 & 34.7 & 31.6 \\
GritLM & 53.2 & \underline{43.3} & \textbf{40.2} & 34.2 & \textbf{33.8} & \underline{36.4} & 47.9 & 41.3 \\
QWEN2  & \underline{62.3} & \textbf{45.6} & 37.1 & 36.8 & 30.9 & 36.3 & \underline{49.5} & \underline{42.6} \\
\bottomrule
\end{tabular}
\end{adjustbox}
\vspace{0.5em}
\caption{Ablation results (nDCG@10, \%) across domains. }
\label{tab:ablation-cleaning}
\end{table*}

\subsubsection{Comparison on Retrieval Metric.}  
We evaluate four cleaning strategies on the BRIGHT dataset:
\begin{itemize}
    % \item \textbf{Raw:} The original BRIGHT chunks without any processing.
    \item \textbf{\clean{} :} retains the two clean agents of \mymethod{} (SafeClean and FastClean) to remove structural noise and boiler-plate; the resulting text is then chunked with BRIGHT’s original heuristic—newline splitting followed by fixed-token truncation.

    \item \textbf{\splits{}:} retains only the split agent from \mymethod{}, which directly segments the original BRIGHT documents into semantically coherent chunks without any prior cleaning.

    \item \textbf{\mymethod{}:} Applies the full \mymethod{} pipeline, as detailed in Section~\ref{sec:method}.

\end{itemize}

Table~\ref{tab:ablation-cleaning} presents the ablation study comparing different cleaning strategies. Several key observations emerge:

\paragraph{\splits{} Achieves Better Retrieval Performance than \mymethod{}.}

The \splits{} configuration performs slightly better than \mymethod{} in terms of average retrieval performance (nDCG@10) across all retrievers. For example, QWEN2 achieves 45.17 with \splits{} compared to 42.63 with the \mymethod{}, and GritLM scores 42.27 vs.\ 41.29 respectively. This indicates that LLM-based semantic segmentation plays a primary role in retrieval effectiveness, while the additional cleaning step contributes marginal improvement under this metric. To further understand their impact on end-to-end RAG performance, we conduct additional ablation studies comparing these two settings in downstream answer generation and LLM-based evaluation.

\begin{table*}[h]
\centering
\footnotesize
\begin{adjustbox}{width=1.0\linewidth}
\begin{tabular}{lccccccccc}
\toprule
\textbf{Method} & Bio. & Earth. & Econ. & Psy. & Rob. & Stack. & Sus. & Avg. & $\Delta$ \\
\midrule
None  & 63.3 & 59.1 & 55.7 & 61.1 & 48.9 & 62.0 & 58.9 & 58.4 & -- \\
\splits{} & 64.8 & \textbf{62.5} & 59.0 & 62.0 & \textbf{50.2} & 62.8 & 59.9 & 60.2 & {\color{red}$\uparrow$83\%} \\
\mymethod{} (Ours)  & \textbf{67.0} & 61.3 & \textbf{59.1} & \textbf{62.9} & 49.4 & \textbf{63.0} & \textbf{60.0} & \textbf{60.4} & {\color{red}$\uparrow$100\%} \\
\bottomrule
\end{tabular}
\end{adjustbox}
\caption{LLM-evaluated QA performance across domains. \mymethod{} and \splits{} both improve over the \textit{None} baseline, with \mymethod{} achieving the highest overall score. $\Delta$ denotes relative improvement.}

\label{tab:end2end-ablation}
\end{table*}

\subsubsection{End-to-End RAG Performance}

To assess the impact of document splitting (\splits{}), we compare three document configurations—\emph{None}, \splits{}, and \mymethod{}—within a unified RAG pipeline for downstream generation.

First, the retriever ranks corpus chunks and returns the top-10 hits by nDCG@10.  
Then, the generator generates an answer conditioned on the user query and the retrieved evidence.  
Lastly, the evaluator assigns a 0–100 score following the relevance-and-coverage rubric.

This three-stage protocol cleanly isolates the contribution of the splitting strategy to downstream RAG quality.

\paragraph{Results.}
The \textit{None} baseline lags behind both \mymethod{} variants across every domain.  
Introducing the \splits{} configuration—splitting without prior cleaning—already lifts retrieval quality appreciably, accounting for roughly four-fifths of the total gain achieved by the full \mymethod{} pipeline.  
Adding the cleaning stage yields a further, but comparatively modest, improvement and secures the top score in most domains.  
These observations confirm that chunk‐level semantic segmentation is the dominant factor driving end-to-end RAG effectiveness, while additional cleaning provides incremental refinement.

\paragraph{Deployment Recommendation.}
We recommend adopting \splits{} as the default preprocessing strategy in production RAG systems. It delivers the majority of performance improvements with minimal token overhead, making it a strong default choice in most scenarios.  
In settings where token budget is less constrained or higher accuracy is essential, the full \mymethod{} pipeline remains a compelling option, offering additional refinement and stronger domain-level consistency.

\subsection{Reasoning Enhancement}

To evaluate the impact of query formulation on downstream reasoning performance, we replace the original queries with reasoning-style rewrites generated by five state-of-the-art LLMs: Claude-3-Opus, Gemini-1.0, GPT-4, Grit, and LLaMA3-70B. These LLM-generated queries are issued to fixed retrievers, and we assess the effectiveness of the retrieved content in supporting reasoning tasks across seven domains.

\begin{figure}[h]
    \centering
    \includegraphics[width=\linewidth]{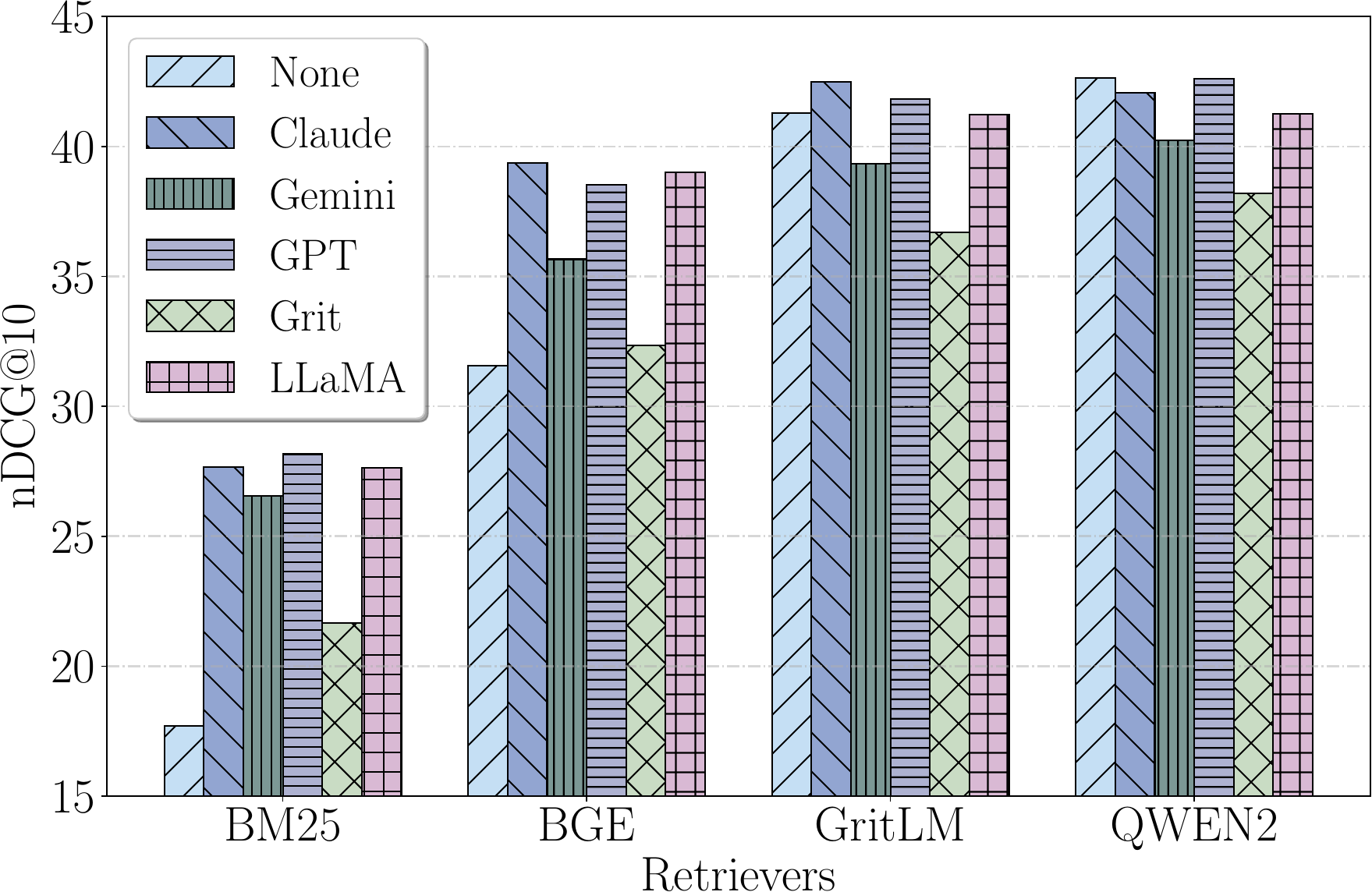}
    \caption{Impact of LLM-rewritten queries on reasoning performance (nDCG@10, \%) across retrievers. See Appendix~\ref{app:reason} for full domain-level results.}
    \label{fig:reasoning}
\end{figure}

\paragraph{Reasoning-style queries benefit weaker retrievers but offer limited gains for stronger models.}
Figure~\ref{fig:reasoning} shows that LLM-generated reasoning queries notably improve retrieval performance for BM25 and BGE, where clearer semantics help guide relevance. However, for high-capacity retrievers like GritLM and QWEN2, performance remains largely unchanged, suggesting these models already capture sufficient semantic cues from standard queries. While Claude and GPT-4 generate consistently strong queries, their impact is retriever-dependent, with diminishing returns on top-tier models.

\subsection{Reranking Enhancement}
\begin{figure}[t]
    \centering
    \includegraphics[width=\linewidth]{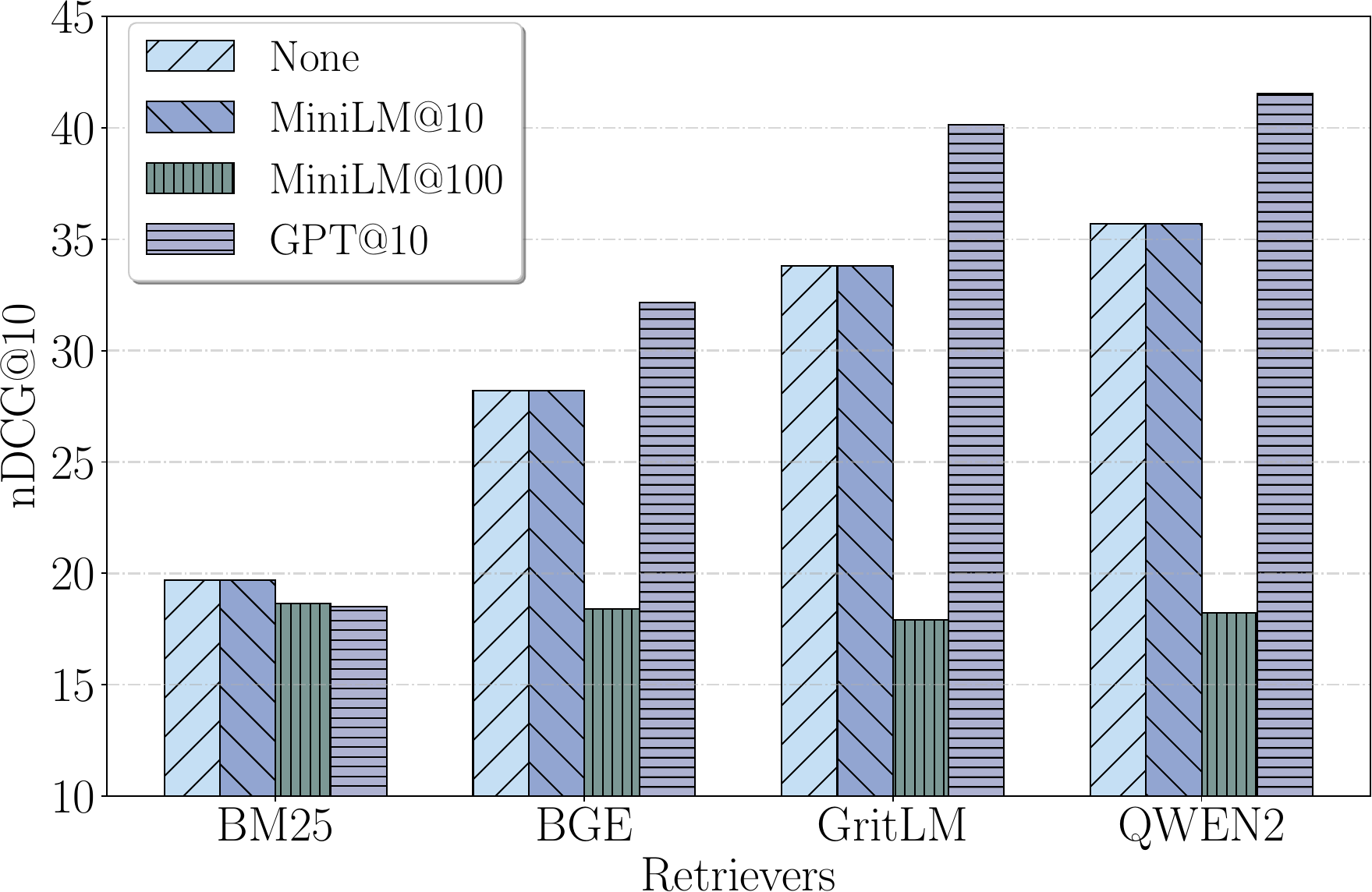}
    \caption{Retrieval metric(nDCG@10, \%) of rerankers applied to top-$k$ results from different retrievers. See Appendix~\ref{app:rerank} for full domain-level results.}
    \label{fig:rerank}
\end{figure}

To evaluate the effectiveness of reranking in improving retrieval quality, we conduct experiments using two rerankers: a lightweight model (MiniLM) and a stronger LLM-based model (GPT-3.5-Turbo). Both are applied to the top-$k$ candidate passages returned by different retrievers. We experiment with two settings for the candidate pool size ($k = 10$ and $k = 100$ for MiniLM, $k = 10$ for GPT-3.5-Turbo) to analyze how reranking performance is affected by retrieval depth. This setup allows us to assess whether rerankers can consistently enhance semantic relevance and how sensitive they are to the retrieval noise introduced by larger candidate sets.
Fig~\ref{fig:rerank} shows the reranking results across different retrievers and candidate set sizes.

\paragraph{Reranking Offers Limited Gains at Small $k$, and Degrades with Large Candidate Pools.}
Reranking with MiniLM@$10$ achieves retrieval performance comparable to the \emph{None} baseline, indicating that lightweight rerankers provide limited benefits when the top-$k$ candidates are already of reasonably high quality. However, increasing the candidate pool to $k=100$ introduces substantial noise, leading MiniLM@$100$ to perform worse than both MiniLM@$10$ and \emph{}{None}, especially with neural retrievers such as GritLM and QWEN2. This highlights the importance of candidate quality before reranking and suggests that rerankers are sensitive to noisy retrieval inputs, particularly when operating over larger passage sets.

\paragraph{Stronger LLM Rerankers Yield the Best Performance.}
GPT-3.5-Turbo@$10$ achieves the highest nDCG@10 across all retrievers, indicating the value of stronger semantic reasoning in reranking. Compared to MiniLM@$10$, GPT-based reranking improves performance by over 5 absolute points for GritLM and QWEN2, highlighting the benefit of LLMs for fine-grained relevance assessment.

\section{Conclusion}

We introduced BRIGHT\textsuperscript{+}, a high-quality refinement of the BRIGHT benchmark tailored for retrieval-augmented reasoning. By identifying and addressing two major flaws—content redundancy and semantic discontinuity—we developed \mymethod{}, an LLM-powered clean-and-split pipeline that produces coherent, retrievable chunks. Our results show that BRIGHT\textsuperscript{+} not only improves retrieval performance across multiple models but also enhances downstream RAG effectiveness. We release the dataset and pipeline to support further research on scalable, semantics-aware data preparation for reasoning-centric applications.

\section*{Limitations}

\mymethod{} improves corpus quality for retrieval-augmented reasoning, but has limitations. Its reliance on large LLMs incurs high computational cost, making it challenging to scale without further optimization. Although SafeClean preserves gold spans and avoids hallucinations, changes to surrounding non-gold regions may still disrupt local semantic continuity, potentially affecting retriever behavior. Moreover, the pipeline depends on BRIGHT-specific supervision, such as span-level annotations, which limits its direct applicability to benchmarks without similar metadata.

\section*{Ethics Statement}
This work focuses on improving the quality of retrieval corpora through LLM-based cleaning and segmentation. All models used in this study are publicly available, and the BRIGHT dataset was released under a research-friendly license. Our proposed pipeline operates on existing textual data and does not involve the collection or generation of personally identifiable information, sensitive content, or human subject data. Nonetheless, we acknowledge that LLM outputs may carry unintended biases or hallucinations. To mitigate this, we adopt conservative prompting strategies that avoid paraphrasing and minimize hallucinated edits. We encourage future users of BRIGHT\textsuperscript{+} and \mymethod{} to exercise similar caution and to conduct additional validation when applying these tools in downstream or real-world settings.

% \section*{Acknowledgements}

\bibliography{acl2020}

\clearpage 
\onecolumn
\appendix  
\section*{Appendix}

\section{Prompt Templates for Different Agents}
\label{app:prompts}

Table~\ref{tab:agent-prompts} presents the full prompt templates used for the three agents—\textsc{SafeClean}, \textsc{FastClean}, and \textsc{Splitter}—in our \mymethod{} pipeline. These prompts are passed to a chat-based LLM in either system or user roles and are designed to elicit consistent, high-precision outputs for content cleaning and segmentation.

\begin{table*}[h]
\centering
\small
\caption{Prompt templates used for each LLM agent in \mymethod{} pipeline.}
\label{tab:agent-prompts}
\begin{tabular}{p{0.1\linewidth} | p{0.83\linewidth}}
\toprule
\textbf{Agent} & \textbf{Prompt Template} \\
\midrule

\textsc{SafeClean} &
\texttt{You are given raw HTML from a Wikipedia page. Your task is to extract the main article content, clean it up, and preserve it as close to the original meaning as possible.} \\
& \texttt{Instructions:} \\
& \texttt{1. Do \textbf{not} rewrite, summarize, or reformat the content.} \\
& \texttt{2. Remove only the following irrelevant elements: navigation bars, sidebars, headers, footers, language links, edit/tool buttons, login links, donation banners, search boxes, and copyright.} \\
& \texttt{3. Keep all meaningful article content, including section titles, paragraphs, lists, citations, and embedded text.} \\
& \texttt{4. If you encounter garbled characters (e.g., ?), try to infer the correct meaning from context; if uncertain, leave unchanged.} \\
& \texttt{5. Output must be valid, clean Markdown only—no extra explanation or commentary.} \\
& \texttt{6. \textbf{Never remove} actual article content or paragraph text.} \\
& \texttt{Raw HTML: \{html\}} \\
\midrule

\textsc{FastClean} &
\texttt{You are given raw text (possibly containing HTML). Extract and return only the main article content.} \\
& \texttt{Rules:} \\
& \texttt{1. Remove irrelevant elements (e.g. navbars, sidebars, headers, footers, edit/login links, banners, ads, search boxes, etc.).} \\
& \texttt{2. If there is no main content, return an empty line.} \\
& \texttt{3. Keep all actual article text intact—no rewriting, summarizing, or reformatting.} \\
& \texttt{4. Only output the cleaned main article content (plain text).} \\
& \texttt{Raw TEXT: \{html\}} \\
\midrule

\textsc{Splitter} &
\texttt{Split the input text into as few coherent, readable chunks as possible, ideally just one. Preserve as much semantic integrity as possible.} \\
& \texttt{Chunking rules:} \\
& \texttt{1. Minimize the number of chunks. Only split when there is a clear and significant shift in topic or structure.} \\
& \texttt{2. Each chunk should be as long as possible while still forming a coherent unit of meaning.} \\
& \texttt{3. Label each chunk in order as "Chunk A:", "Chunk B:", "Chunk C:", etc.} \\
& \texttt{4. If the input contains tables or structured data, summarize them in natural language and include as part of a chunk.} \\
& \texttt{5. Do not modify the original text content except for structured data. You may delete content at the beginning or end, but preserve original wording and structure elsewhere.} \\
& \texttt{Example output:} \\
& \texttt{Chunk A:\newline This entire chunk stays together because it maintains a consistent topic and logical flow.} \\
& \texttt{Even if it spans several sentences or paragraphs, it is not split unless the topic clearly changes.} \\
\bottomrule
\end{tabular}
\end{table*}

\clearpage 
\section{Reasoning Enhancement Results}
\label{app:reason}
To assess the downstream reasoning capabilities of different retrievers under various query formulations, we conduct an evaluation using LLM-rewritten queries across seven domains in BRIGHT+. Table~\ref{tab:reasoning-results-percent} summarizes the results.

Specifically, we compare performance under six different query-generation settings: the original BRIGHT\textsuperscript{+} queries (None), identical to those in BRIGHT), and rewritten versions produced by five leading LLMs—Claude-3-Opus, Gemini-1.0, GPT-4, Grit, and LLaMA-3. All query rewrites are provided as part of the BRIGHT benchmark suite, enabling controlled comparisons without introducing prompt engineering or LLM inference noise. For each setting, the same document corpus and retrievers are used, isolating the impact of query reformulation.

Across the board, we observe that LLM-generated reasoning-style queries lead to significant gains in nDCG@10, particularly for weaker retrievers such as BM25 and BGE. For instance, BM25 improves from 22.7\% to 45.6\% on Biology when using Gemini-generated queries, while BGE sees consistent boosts across all domains with Claude and GPT queries.

Interestingly, stronger dense retrievers like GritLM and QWEN2 benefit less dramatically from query rewriting, as they already capture much of the underlying semantics. However, even in those cases, high-quality prompts (e.g., from GPT-4 or Claude) yield modest but measurable improvements, suggesting some complementarity between retriever embeddings and prompt expressiveness.

This experiment highlights that while model quality is important, the phrasing of the input query can be an equally powerful lever for improving retrieval precision in reasoning-intensive tasks. Incorporating LLMs into the query preprocessing stage is thus a lightweight yet highly effective enhancement for real-world RAG applications.

\begin{table*}[h]
\caption{Reasoning task results across models and retrievers}
\centering
\scriptsize
\begin{tabular}{l|ccccccc}
\toprule
\textbf{Retriever} & \textbf{Biology} & \textbf{Earth Science} & \textbf{Economics} & \textbf{Psychology} & \textbf{Robotics} & \textbf{Stackoverflow} & \textbf{Sustainable Living} \\
\midrule
\multicolumn{8}{c}{\textbf{None}} \\
\midrule
BM25   & 22.70 & 22.66 & 16.06 & 12.72 & 21.69 & 11.61 & 16.37 \\
BGE    & 36.05 & 33.63 & 31.01 & 25.39 & 27.18 & 32.89 & 34.74 \\
GritLM & 53.23 & 43.25 & 40.18 & 34.24 & 33.82 & 36.38 & 47.92 \\
Qwen2  & 62.25 & 45.59 & 37.08 & 36.80 & 30.88 & 36.44 & 49.49 \\
\midrule
\multicolumn{8}{c}{\textbf{Claude-3-Opus}} \\
\midrule
BM25   & 44.63 & 34.41 & 25.01 & 26.32 & 20.41 & 19.77 & 23.14 \\
BGE    & 55.24 & 40.56 & 37.44 & 31.28 & 28.30 & 39.19 & 43.53 \\
GritLM & 54.65 & 45.02 & 38.99 & 34.88 & 35.00 & 41.99 & 46.78 \\
QWEN2  & 60.27 & 43.30 & 34.93 & 38.26 & 32.91 & 37.66 & 47.04 \\
\midrule
\multicolumn{8}{c}{\textbf{Gemini-1.0}} \\
\midrule
BM25   & 45.60 & 31.65 & 23.99 & 25.92 & 21.85 & 18.72 & 18.09 \\
BGE    & 52.69 & 38.58 & 33.26 & 29.20 & 24.26 & 32.71 & 38.91 \\
GritLM & 53.89 & 42.12 & 37.46 & 31.83 & 30.80 & 36.62 & 42.55 \\
QWEN2  & 59.76 & 43.08 & 29.66 & 34.44 & 33.95 & 35.90 & 44.87 \\
\midrule
\multicolumn{8}{c}{\textbf{GPT-4}} \\
\midrule
BM25   & 44.04 & 36.42 & 24.12 & 27.23 & 23.59 & 21.53 & 20.34 \\
BGE    & 53.66 & 41.08 & 36.54 & 31.49 & 25.25 & 40.77 & 40.91 \\
GritLM & 57.21 & 45.41 & 35.27 & 33.72 & 34.97 & 40.27 & 45.89 \\
QWEN2  & 59.10 & 46.50 & 34.44 & 36.28 & 33.61 & 39.82 & 48.41 \\
\midrule
\multicolumn{8}{c}{\textbf{Grit}} \\
\midrule
BM25   & 31.12 & 24.16 & 21.90 & 21.11 & 19.59 & 16.26 & 17.40 \\
BGE    & 41.80 & 32.48 & 33.37 & 28.09 & 22.19 & 31.25 & 37.27 \\
GritLM & 47.74 & 39.70 & 36.71 & 30.18 & 28.34 & 33.56 & 40.55 \\
QWEN2  & 52.22 & 39.11 & 31.88 & 35.26 & 30.40 & 35.03 & 43.40 \\
\midrule
\multicolumn{8}{c}{\textbf{LLaMA}} \\
\midrule
BM25   & 45.81 & 33.87 & 25.20 & 25.97 & 20.75 & 19.37 & 22.47 \\
BGE    & 53.70 & 40.07 & 36.07 & 30.83 & 29.85 & 38.34 & 44.10 \\
GritLM & 54.57 & 45.11 & 38.30 & 33.18 & 33.03 & 39.37 & 44.95 \\
QWEN2  & 58.30 & 43.64 & 34.11 & 38.38 & 33.83 & 35.15 & 45.46 \\
\bottomrule
\end{tabular}
\label{tab:reasoning-results-percent}
\end{table*}

\clearpage
\section{Reranking Enhancement Results}
\label{app:rerank}

We further investigate the effect of reranking in the BRIGHT\textsuperscript{+} setting by evaluating a second-stage reranker across a range of retrievers and query rewriting conditions. Table~\ref{tab:rerank-results} presents nDCG@10 scores across seven domains.

In this experiment, each retriever produces a candidate list of top-$k$ passages, which are optionally re-scored by a reranker. We consider three reranking conditions:

\begin{itemize}
    \item \textbf{None}: No reranker is used; evaluation is based on the original ranking from the base retriever.
    \item \textbf{MiniLM}: We apply the pretrained cross-encoder \texttt{cross-encoder/ms-marco-MiniLM-L-12-v2} from HuggingFace, a lightweight but effective model trained on MS MARCO.
    \item \textbf{GPT}: We use \texttt{gpt-3.5-turbo} (via OpenAI API) as a reranker, implemented by prompting the model to rank candidate passages in terms of relevance to the input query.
\end{itemize}

We evaluate both low-recall ($k=10$) and high-recall ($k=100$) settings, using the same query variants and document corpus for all runs.
Several key observations emerge:

\begin{itemize}
    \item With \textbf{MiniLM}, reranking tends to mirror the base retriever unless the initial candidates are already strong. Gains are minimal, especially at $k=10$.
    
    \item In contrast, \textbf{GPT-3.5-turbo} shows significant improvements, particularly when reranking outputs from strong retrievers like QWEN2. For example, QWEN2 reranked by GPT at $k=10$ reaches 59.93\% on Biology.
    
    \item At $k=100$, MiniLM reranking often degrades performance due to over-saturation, whereas GPT reranking remains robust, highlighting its superior semantic discrimination.
    
    \item Notably, reranking even benefits classical BM25, though the ceiling is naturally lower.
\end{itemize}

These results suggest that while lightweight rerankers like MiniLM are practical for latency-sensitive applications, high-capacity rerankers such as GPT-3.5 can offer substantial accuracy gains in settings where semantic depth and reasoning are critical.

\begin{table*}[h]
\centering
\caption{Reranking performance (nDCG@10, \%) across 7 BRIGHT domains with different retrievers and top-$k$ settings.}
\label{tab:rerank-results}
\footnotesize
\begin{adjustbox}{max width=\textwidth}
\begin{tabular}{clccccccc}
\toprule
\textbf{Top-$k$} & \textbf{Retriever} & Biology & Earth Science & Economics & Psychology & Robotics & Stackoverflow & Sustainable Living \\
\midrule
\multicolumn{9}{c}{\textbf{None}} \\
\midrule
\multirow{4}{*}{--}  
  & BM25   & 22.41 & 24.27 & 14.27 & 15.84 & 24.91 & 16.53 & 19.67 \\
  & BGE    & 32.12 & 30.98 & 27.76 & 22.01 & 23.16 & 30.31 & 31.06 \\
  & GritLM & 39.98 & 37.87 & 30.59 & 28.37 & 28.95 & 34.49 & 36.36 \\
  & QWEN2  & 48.92 & 38.97 & 30.53 & 30.37 & 26.53 & 32.34 & 42.29 \\
\midrule
\multicolumn{9}{c}{\textbf{MiniLM}} \\
\midrule
\multirow{4}{*}{k=10}  
  & BM25   & 22.41 & 24.27 & 14.27 & 15.84 & 24.91 & 16.53 & 19.67 \\
  & BGE    & 32.12 & 30.98 & 27.76 & 22.01 & 23.16 & 30.31 & 31.06 \\
  & GritLM & 39.98 & 37.87 & 30.59 & 28.37 & 28.95 & 34.49 & 36.36 \\
  & QWEN2  & 48.92 & 38.97 & 30.53 & 30.37 & 26.53 & 32.34 & 42.29 \\
\midrule
\multirow{4}{*}{k=100} 
  & BM25   & 19.37 & 21.77 & 16.53 & 15.74 & 14.14 & 19.96 & 23.01 \\
  & BGE    & 19.62 & 23.19 & 16.78 & 15.51 & 13.26 & 19.23 & 21.26 \\
  & GritLM & 16.48 & 23.58 & 16.38 & 16.18 & 14.82 & 17.89 & 20.08 \\
  & QWEN2  & 17.42 & 23.85 & 16.59 & 14.44 & 15.28 & 17.64 & 22.33 \\
\midrule
\multicolumn{9}{c}{\textbf{GPT}} \\
\midrule
\multirow{4}{*}{k=10} 
  & BM25   & 22.91 & 22.90 & 16.03 & 12.65 & 24.95 & 13.01 & 17.11 \\
  & BGE    & 36.86 & 33.32 & 29.87 & 26.03 & 28.51 & 33.73 & 36.86 \\
  & GritLM & 53.27 & 43.08 & 39.57 & 34.82 & 33.14 & 32.85 & 44.24 \\
  & QWEN2  & 59.93 & 43.32 & 34.40 & 33.36 & 33.20 & 35.15 & 51.39 \\ 
\bottomrule
\end{tabular}
\end{adjustbox}
\end{table*}

\end{document}